\documentclass[letterpaper, 10 pt, conference]{ieeeconf}

\IEEEoverridecommandlockouts             
                                                          
\usepackage{cite}
\usepackage{amsmath,amssymb,amsfonts}
\usepackage{algorithmic}
\usepackage{graphicx}
\usepackage{textcomp}
\usepackage{xcolor}
\usepackage{booktabs}
\usepackage{multirow}
\usepackage{hyperref}
\usepackage{float}
\def\BibTeX{{\rm B\kern-.05em{\sc i\kern-.025em b}\kern-.08em
    T\kern-.1667em\lower.7ex\hbox{E}\kern-.125emX}}

\title{\LARGE \bf
Vibration Suppression in Collaborative Flexible Payload Manipulation Using Passive Force Control
}

\author{Alaa Abderrahim$^{1,4}$, Antonio Rosales$^{1*}$, Ferdinando Milella,$^{2}$ Markku Suomalainen$^{1}$, and Shuai Li${^{1,3}}$ 
\thanks{This project has received funding from Research Council of Finland project UNITE: Unifying and optimizing representations in robot learning. This work has been carried out within the framework of the EUROfusion Consortium, funded by the European Union via the Euratom Research and Training Programme (Grant Agreement No 101052200 — EUROfusion). Views and opinions expressed are however those of the author(s) only and do not necessarily reflect those of the European Union or the European Commission. Neither the European Union nor the European Commission can be held responsible for them.}
\thanks{ $^{1}$ A. Abderrahim, A. Rosales, M. Suomalainen, and S. Li are with VTT Technical Research Centre of Finland Ltd, Oulu, Finland.}
\thanks{$^{2}$ Ferdinando Milella is with the UK Atomic Energy Authority, Oxfordshire, England, United Kingdom.}
\thanks{$^{3}$Shuai Li is with the Faculty of Information Technology and Electrical Engineering, University of Oulu, Finland.}
\thanks{$^4$ A. Abderrahim is an electrical engineering student at the, RP Technical University of Kaiserslautern, Germany}
\thanks{*Corresponding author: {\tt\small antonio.rosales@vtt.fi}}
    }
     
\begin{document}

\maketitle
\thispagestyle{empty}
\pagestyle{empty}

\begin{abstract}
In large and heavy structures, vibrations arise during motion, posing significant challenges for precise manipulation. To accomplish the desired motion, control algorithms must effectively suppress these structural vibrations. In cutting-edge projects, such as remote maintenance of future fusion energy reactors (tokamaks), the manipulation of this type of structure is defined as a crucial task. This paper presents a control strategy to suppress transverse vibrations in flexible payloads during motion using a collaborative payload manipulation approach. Two different industrial robot arms are arranged in a leader–follower configuration for the manipulation strategy. The leader robot guides the motion with shaped velocity commands, while the follower robot ensures compliance with the estimated external forces applied by the leader on the payload through an admittance controller. Unlike existing methods, the proposed approach enables collaborative manipulation of heavier and larger flexible objects, addressing additional challenges such as vibration suppression and heterogeneous robot specifications. The dynamics of the leader–follower–payload system are modeled using an equivalent mass-spring-damper model, and it is shown that, with appropriate admittance parameters, the total energy of the system is passively dissipated. A stability proof is also provided. Numerical simulations validate the proposed method, and experimental results (see online video \cite{alaa2025} ) demonstrate its effectiveness.
\end{abstract}


\section{Introduction}
Massive and considerably large structures that present flexibility due to their size represent a challenge in the context of flexible object manipulation. The manipulation of these objects is essential for ongoing and future cutting-edge projects, such as performing on-orbit services \cite{KawaiIET2021} and remote maintenance of fusion energy reactors \cite{eudemo2022}.

The control of robots that manipulate and handle flexible materials is not straightforward. Suppressing/damping the oscillations that appear during the manipulation of the flexible object is crucial since these oscillations may damage the object itself and produce disturbances on the robot. The controller must be capable of accomplishing the desired motion task while suppressing/damping oscillations that emerge on the object \cite{Cherubini2020}.

Research on the manipulation of flexible objects has mostly been focused on relatively light and small objects, such as flexible beams and metal sheets. In the case of massive structure manipulation, research focuses on proof-of-concepts using simulations \cite{TIKKA2025114989} and industrial robots for experimental tests \cite{HERSCHMANN2024114392} since there are no real-scale experimental setups. 

\begin{figure}
  \begin{center}
    \includegraphics[angle=0, width=8.5cm]{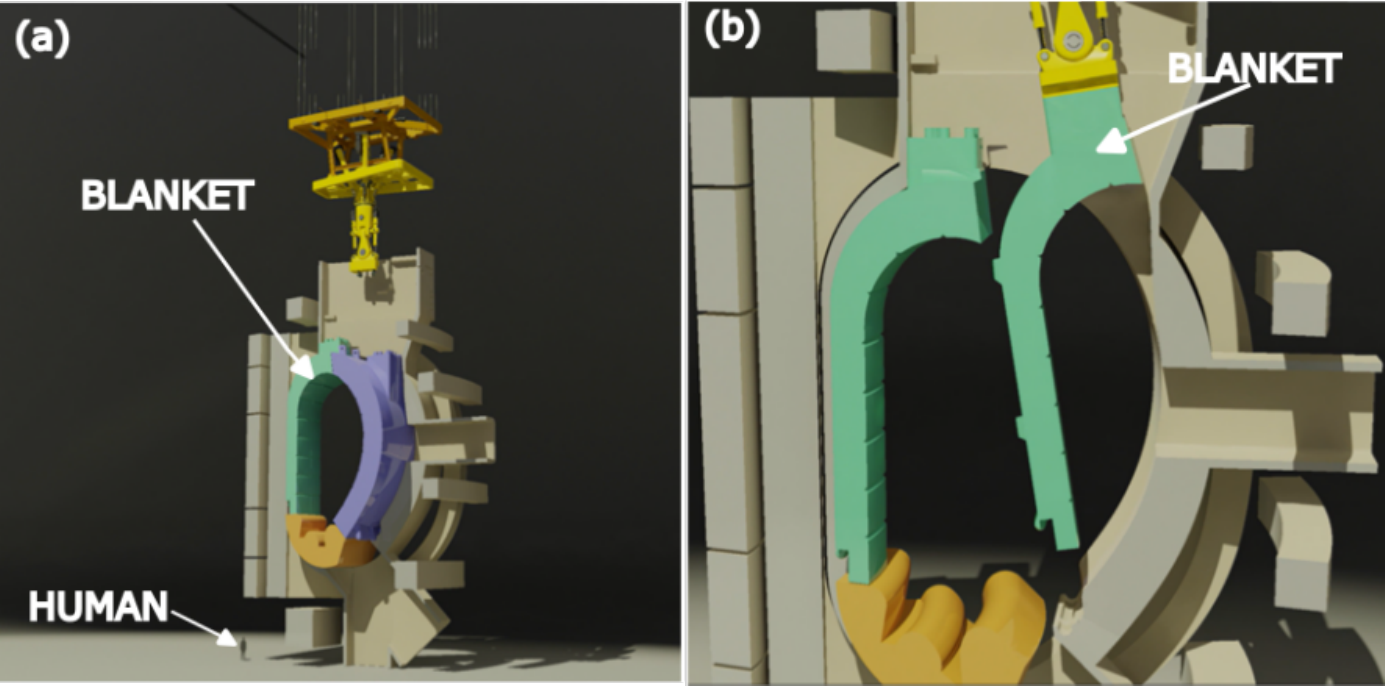}
  \end{center}
    \caption[]{ (a) Model simulation of the breeding blanket in the conceptual DEMOnstration power plant, and (b) its removal process.}
    \label{fig:intro}
\end{figure}

In this paper, we study the manipulation of flexible objects using industrial robots. The aim is to investigate a solution for the manipulation of massive and large objects, such as in-vessel remote handling of breeding blankets, a critical challenge in the Demonstration Fusion Power Plant (DEMO) project \cite{eudemo2022}. Figure \ref{fig:intro} presents a simulation model of the DEMO plant. The blanket segments to be manipulated are longer than 10 meters and weigh more than 80 tonnes, see its size compared with a human in Figure \ref{fig:intro}.(a). The blankets should be removed after several years of use, when they have already lost their rigidity. Also, the manipulation of the structure is not in free space, see Figure \ref{fig:intro}.(b), increasing the difficulty of the task. It is envisaged that clearance around blanket segments in future tokamaks will be very small (below 1cm), which makes oscillation damping even more crucial. 

Recently, a Two-port Mover (TPM) solution has been proposed to overcome the difficulties of blanket segment handling in DEMO \cite{TIKKA2025114989}. However, the proposed solution (based on a system of cables and pulleys) does not provide the required capability of damping oscillations during the extraction manoeuvre. Therefore, alternative solutions based on the deployment of two fully actuated robotic systems at the upper and lower ports of the tokamak vessel are currently being considered. Although the concept design of a fully actuated TPM is not yet fully defined, the related control problem can be essentially represented as a two-robot cooperative handling of a flexible payload. For a preliminary assessment of the best control strategy to develop for a TPM, we have considered the simple case of two robotic arms with differing capabilities that manipulate a flexible bar, as shown in Figure \ref{fig:fig1}. One heavy-duty arm prioritizes strength and executes high-force tasks, such as lifting, and a second lightweight arm executes high-precision tasks like positioning/alignment. This strategy aligns with the fully actuated TPM concept under investigation. The two robot arms are coordinated using a leader-follower approach based on admittance control.

The suppression of oscillations/vibrations on flexible structures during their manipulation with robot arms is mainly addressed by the following two approaches. One approach utilizes online feedback from the oscillations/vibrations of the structure to compute the control input of the robot. Feedback signals such as deflection, rotation, strain, and shear forces are used. One example is adaptive control with online deflection measurements to actively suppress oscillations by adjusting the control gains \cite{LEE2001951}. One more example is the work presented in \cite{TAVASOLI2009212}, where a linear observer is used to estimate information about the vibration, and these estimated variables are used to compute the control law. The second approach to suppress oscillations/vibrations on flexible structures does not utilize any online information about the vibrations/oscillations of the structure to compute the control input. Input shaping is one of these methods, as it involves designing an input command that avoids exciting the resonance frequency of the flexible body \cite{kotaniemi2025data}. Another example is the PD (Proportional-Derivative) controller, which regulates the position of the flexible beam without requiring information about the vibration \cite{LiuSunTAC2000}. Despite the vast literature addressing the oscillation suppression of flexible structures using two-robot arms, few of the presented approaches are validated experimentally.

The integration of admittance and impedance control into leader-follower schemes has been mainly used to manipulate rigid objects. For example, a leader-follower approach to manipulate rigid objects was proposed in \cite{MingheIROS14}. In \cite{ZhaoICCCR2023}, the authors presented a leader-follower scheme focused on a nuclear industry study. An adaptive impedance control approach for leader-follower robotic systems is presented in \cite{SATVATI2025101274}. The manipulation of flexible objects is rarely addressed via leader-follower approaches with admittance control.

A two-robot collaborative method for manipulating flexible structures that does not use online information about the structure vibrations is presented in this paper. The motivation to propose a solution without vibration feedback for computing the control input is the difficulty of placing vibration/deflection sensors on structures like breeding blankets, where sensor placement is not easy due to gamma radiation \cite{HERSCHMANN2024114392}. We present a leader-follower scheme, where the leader element is an industrial robot that lifts/carries the flexible payload from one side, and the follower element is a lightweight robot holding the payload from the other side. Then, the industrial robot leads the manipulation, moving the payload, while the lightweight robot follows the movement. Compliance is added to the lightweight robot via an admittance control that maps the force estimated/measured at the tip of the industrial robot to velocity commands sent to the lightweight robot. The lightweight robot admittance is mainly damping to suppress oscillations on the flexible object. Simulations of the collaborative method using Matlab/Simulink are presented. Experimental tests using a KUKA-quantek robot collaborating with a KUKA-Iiwa robot to move a flexible sheet demonstrate the effectiveness of the proposed scheme.

The novelty of this paper lies firstly in the use of a leader-follower scheme with admittance to manipulate flexible objects. Also, compared with most of the leader-follower robot arm schemes, we are considering robots with different load capabilities, allowing the manipulation of heavy objects. Furthermore, a distinctive contribution of this study is that, unlike much of the existing literature, which remains largely simulation-based, we provide experimental validation of the proposed method on a physical setup, demonstrating its practical feasibility and effectiveness in real robotic applications.


\section{Description of the collaborative system}\label{sec:descip}
\begin{figure}[H]
  \begin{center}
    \includegraphics[angle=0, width=5cm]{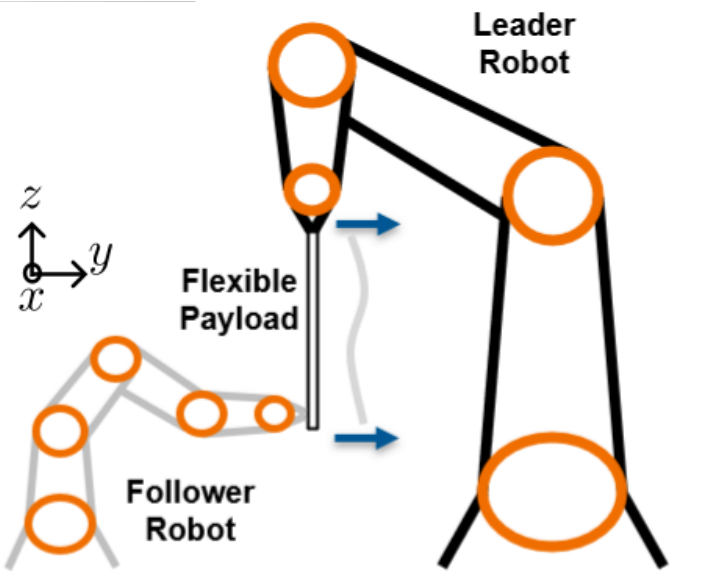}
  \end{center}
    \caption[]{Two-robot collaborative system }
    \label{fig:fig1}
\end{figure}

The two-robot collaborative system consists of one heavy-duty robot and one lightweight robot handling a flexible payload. The payload is a flexible rectangular structure. Only translational vibrations on the structure are considered, see Figure \ref{fig:fig1}. The task to be performed is to collaboratively move the flexible structure in the horizontal direction ($y$-direction), avoiding oscillations on the structure. The goal is to develop a collaborative control scheme for the robots to achieve the motion task.

The proposed collaborative scheme works as follows. The heavy-duty robot leads the collaboration, receiving a velocity/position command in the $y$-direction. When the heavy-duty robot begins the motion, an interaction force generated by the physical attachment with the flexible structure and the lightweight robot is exerted on its end-effector. This interaction force is measured/estimated and sent to the lightweight robot via an admittance controller. Then, the lightweight robot follows the movement of the structure and the leader robot. The details of the design of the collaborative control scheme are presented in the next section \ref{sec:VibSupre}.

\section{Vibration suppression via Passive Force control}\label{sec:VibSupre}
\begin{figure}
  \begin{center}
    \includegraphics[angle=0, width=7cm]{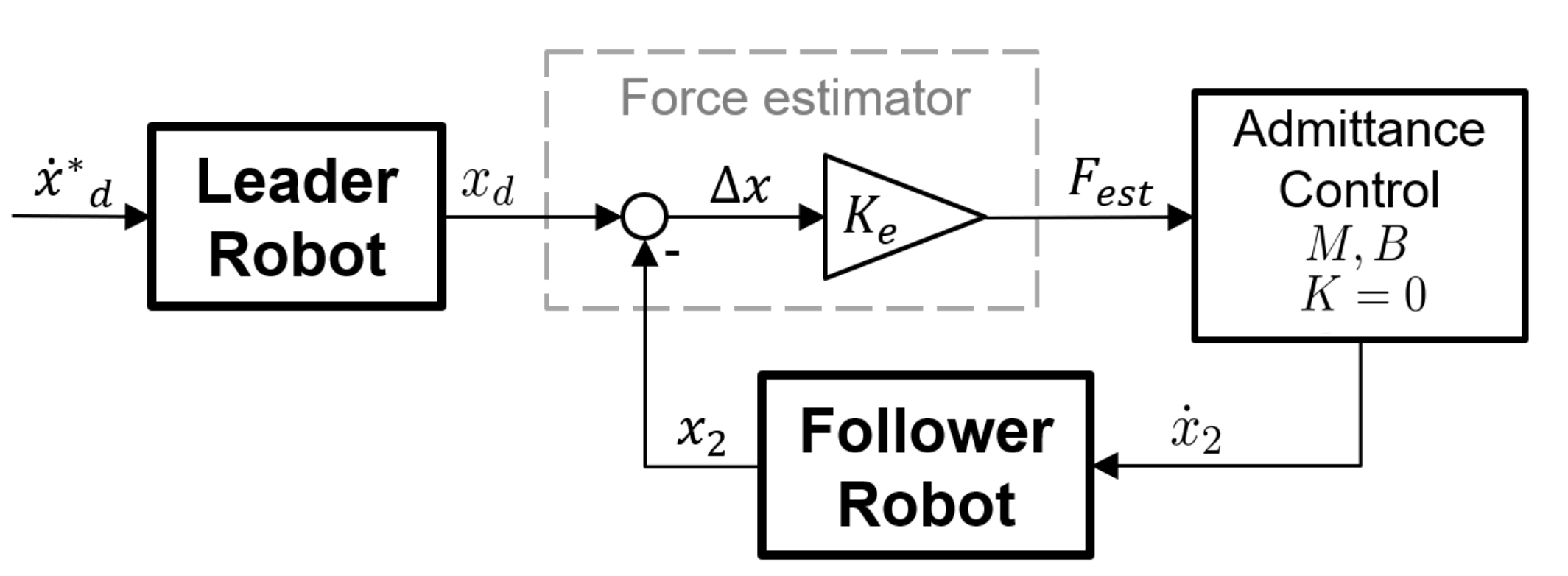}
  \end{center}
    \caption[]{ Block diagram of collaborative scheme. $\dot{x}^{*}_d$ is the leader's velocity command, $\dot{x}_{2}$ is the follower's velocity command, and the estimated interaction force is $F_{est}$}
    \label{fig:block}
\end{figure}

A block diagram of the proposed collaborative scheme is presented in \autoref{fig:block}. The leader robot tracks a reference trajectory $\dot{x}^{*}_d$ provided by either an external operator or a path-planning algorithm while grasping, lifting, and pulling the payload from the top side. The leader should behave as if it was carrying the payload alone. On the other hand, the follower robot grasps the payload from the bottom side, and it dynamically adjusts its motion to remain compliant with the leader. Admittance control is used in the follower robot to have compliance. In the classical formulation of admittance control, the input to the admittance block is an external force measured by force/torque sensors, and the output is a velocity command. Here, to account for the vibrations of the flexible body and noise, we adopt a variation inspired by cross-coupling control \cite{sun2006synchronous}. Instead of directly using force measurements, we use an elastic model considering the position error between the leader and follower robots, multiply it by a stiffness gain, and treat this signal as an estimated force $F_{est}$. This estimated force is fed to the follower admittance controller, whose parameters are chosen to have only mass-damping dynamics without spring effect. Then, the follower generates compliant velocity commands that drive its motion to reduce the position error following the leader more effectively.

\subsection{Passivity Analysis}
\begin{figure}
  \begin{center}
    \includegraphics[angle=0, width=7cm]{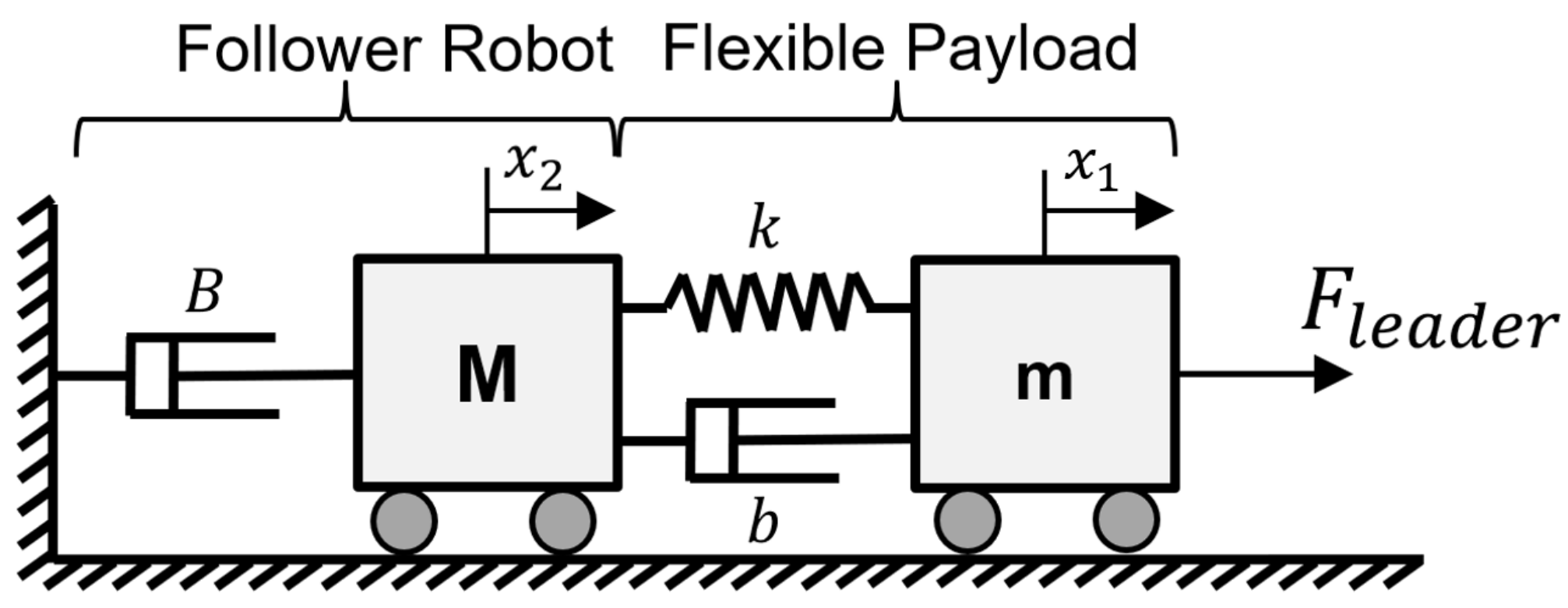}
  \end{center}
    \caption[]{Mass-spring damper system representation of the follower-payload system with force applied from the leader robot}
    \label{fig:MBK}
\end{figure}

In this subsection, we present a total energy analysis of the follower-payload system to demonstrate how the proposed control strategy suppresses vibrations. The follower robot is modeled as a lumped mass–damper system, and the flexible payload is represented as a mass–spring–damper system. An external force \(F_{\text{leader}}\) is applied to the system by the leader robot, see \autoref{fig:MBK}.

Consider the mass-spring damper representation in \autoref{fig:MBK}, the dynamics of the system are given by:
\begin{align}
M \ddot{x}_2 &= k(x_1 - x_2) + b(\dot{x}_1 - \dot{x}_2) - B\dot{x}_2, \label{eq:dyn1} \\
m \ddot{x}_1 &= F_{\text{leader}} - k(x_1 - x_2) - b(\dot{x}_1 - \dot{x}_2), \label{eq:dyn2}
\end{align}
where \(M\), \(x_2\), and \(\dot{x}_2\) are the mass, displacement, and velocity of the follower robot, respectively. The mass, displacement, and velocity of the payload are \(m\), \(x_1\), and \(\dot{x}_1\), respectively. The constants \(B\) and \(b\) are the damping coefficients of the follower and the payload, respectively, while \(k\) is the spring constant of the payload.  

The total mechanical energy $E_T$ of the system is composed of the kinetic energy $K_f$ of the follower robot and the payload $K_p$, which represents their motion, as well as the elastic potential energy stored in the payload spring $U_p$, which reflects the vibrations.
\begin{equation}
E_T = K_f + K_p + U_p 
= \tfrac{1}{2}M\dot{x}_2^2 + \tfrac{1}{2}m\dot{x}_1^2 + \tfrac{1}{2}k e^2,
\end{equation}
where \(e = x_1 - x_2\) is the relative displacement error between payload and follower.  

Taking the time derivative of the total energy:
\begin{equation}
\dot{E}_T = M\dot{x}_2\ddot{x}_2 + m\dot{x}_1\ddot{x}_1 + k e \dot{e}.
\end{equation}
By substituting the dynamics equations \eqref{eq:dyn1} and \eqref{eq:dyn2} into the above expression, we obtain:
\begin{equation}
\dot{E}_T = \dot{x}_2 \big(k e + b\dot{e} - B\dot{x}_2\big) 
+ \dot{x}_1\big(F_{\text{leader}} - k e - b\dot{e}\big) 
+ k e \dot{e}.
\end{equation}

After simplification, the total energy rate becomes:
\begin{equation}
\dot{E}_T = \dot{x}_1 F_{\text{leader}} - B\dot{x}_2^2 - b\dot{e}^2.
\end{equation}

For every value for \(\dot{x}_2\) and \(\dot{e}\), the terms \(- B\dot{x}_2^2 - b\dot{e}^2\) are negative, corresponding to the energy dissipated by the damping elements of the follower and the payload. 

The rate of change of the total energy satisfies  
\begin{equation}
\dot{E}_T(t) = \dot{x}_1(t) F_{\text{leader}}(t) - B \dot{x}_2^2(t) - b \dot{e}^2(t).
\end{equation}
Integration of this relation from $t=0$ to $t=T$ gives
\begin{equation}
\begin{split}
E_T(T) = E_T(0) 
 &+ \int_0^T \dot{x}_1(t)F_{\text{leader}}(t)\,dt \\
 &- \int_0^T \big(B \dot{x}_2^2(t) + b \dot{e}^2(t)\big)\,dt .
\end{split}
\end{equation}
This expression shows that the total mechanical energy at time $T$ is equal to the initial stored energy, plus the energy supplied by the leader through the term $\int_0^T \dot{x}_1 F_{\text{leader}}\,dt$, minus the energy dissipated by the damping elements. In particular, if $F_{\text{leader}} \equiv 0$, the total energy decreases monotonically, ensuring asymptotic decay of vibrations.  

Furthermore, since
\begin{equation}
\int_0^T \dot{x}_1(t) F_{\text{leader}}(t)\,dt \;\geq\; - E_T(0),
\end{equation}
the system satisfies the passivity inequality with input $F_{\text{leader}}$ and output $\dot{x}_1$ \cite{brogliato2007dissipative}. Physically, this means that the energy which can be extracted from the system is always bounded by the initial stored energy. The follower--payload dynamics therefore constitute a passive mapping from the leader's force input to the payload's velocity output, ensuring robustness against excitation and providing a natural mechanism for vibration suppression.



\subsection{Stability Analysis of the Scheme}
From the equation dynamics \eqref{eq:dyn1} and \eqref{eq:dyn2} we choose the state $x=\left[\begin{matrix}x_1 & x_2 & \dot x_1 & \dot x_2\end{matrix}\right]^\top$
the force as input $u=F_{leader}$, and the velocities as output $y=\begin{bmatrix}\dot x_1 & \dot x_2\end{bmatrix}^\top$, which yields the state--space model 
\[
\dot x = \underbrace{\left[\begin{smallmatrix}
0 & 0 & 1 & 0 \\
0 & 0 & 0 & 1 \\
-\frac{k}{m} & \frac{k}{m} & -\frac{b}{m} & \frac{b}{m} \\
\frac{k}{M} & -\frac{k}{M} & \frac{b}{M} & -\frac{b+B}{M}
\end{smallmatrix}\right]}_{A} x + \underbrace{\left[\begin{smallmatrix}
0 \\
0 \\
\tfrac{1}{m} \\
0
\end{smallmatrix}\right]}_B u, 
\]
\[
y = \underbrace{\left[\begin{smallmatrix}
0 & 0 & 1 & 0 \\
0 & 0 & 0 & 1
\end{smallmatrix}\right]}_C x,
\]


The transfer functions are obtained as $G(s)=C(sI-A)^{-1}B+D=\frac{1}{\Delta(s)}
\left[\begin{smallmatrix}
M s^2+(B+b)s+k \\
b s+k
\end{smallmatrix}\right]$,where $\Delta(s)=a_3 s^3+a_2 s^2+a_1 s+a_0$, with $a_3 = M m$, $a_2 = B m + M b + b m$, $a_1 = B b + (M+m)k$, and $a_0 = B k$.

All coefficients are strictly positive for $m,M,k>0$, $b\ge 0$, and $B>0$. 
According to the Routh-Hurwitz criterion, a cubic polynomial is Hurwitz 
(i.e., all roots lie in the open left half-plane) if $a_3,a_2,a_1,a_0>0$ and
$a_2 a_1 > a_3 a_0$. Here, this inequality expands to
\[
\bigl(\underline{Bm}+Mb+bm\bigr)\,\bigl(Bb+\underline{Mk}+mk\bigr) - (\underline{Mm})(\underline{Bk}) > 0,
\]
where the underlined terms are substracted.  
Since all the remaining terms are strictly positive, the inequality is always satisfied.
which is always satisfied since every term is positive.  
Therefore, $\Delta(s)$ is Hurwitz, and all poles of $G(s)$ lie in the open left half-plane. 
It follows that the input--output map $F_{leader} \mapsto (\dot x_1,\dot x_2)$ is \textbf{BIBO stable}.

\subsection{Design of the leader control input}
\begin{figure}
  \begin{center}
    \includegraphics[angle=0, width=6cm]{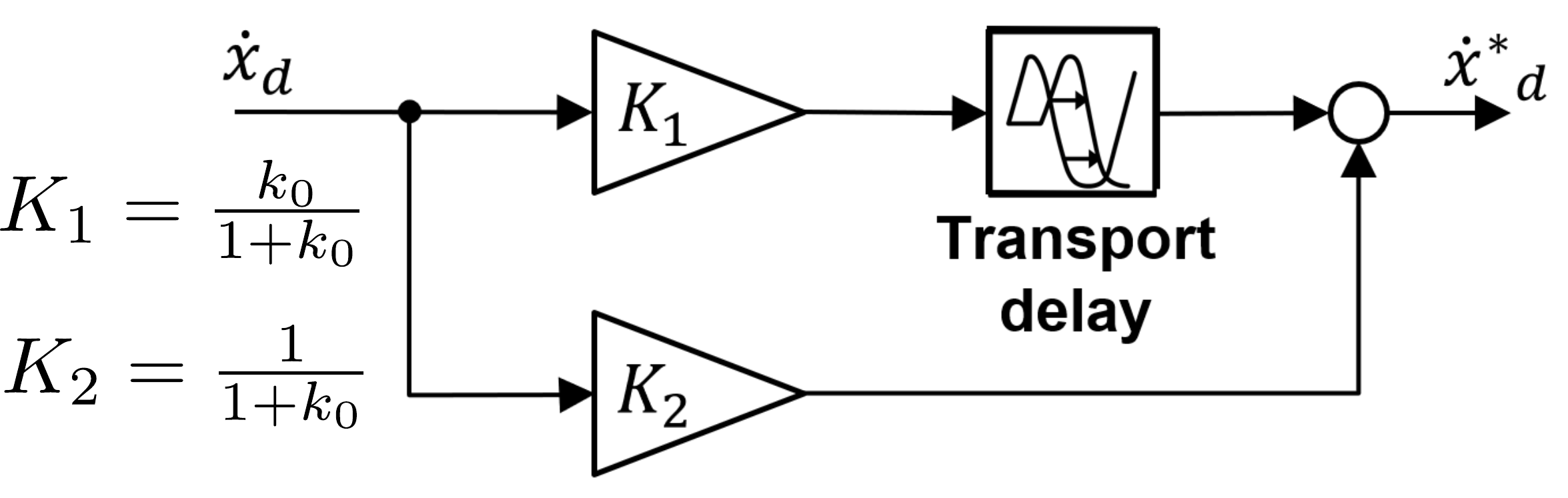}
  \end{center}
    \caption[] {Block diagram of input-shaping. }
    \label{fig:IS}
\end{figure}

The leader robot feeds energy to the collaborative system. Thus, a poorly planned reference input might introduce more undesired residual vibrations, decreasing accuracy and compromising the structural integrity \cite{khan2020sliding}. Therefore, we decided to use input shaping to compute the input of the leader. Input-shaping modifies the reference motion/velocity command signal for a dynamic system by effectively splitting it into multiple sub-signals and introducing specific time delays between them, based on the half-period of the unwanted frequency\cite{singer1990preshaping}, such that it avoids introducing unwanted oscillations in the system. 
It is important to note that the proposed input shaping method is purely feed-forward. Even though external sensors are required to find the natural frequency of the robot manipulator, it does not require constant feedback. Moreover, input shaping only requires the ”model” of the system in the weakest sense, namely, the natural frequency of the system is sufficient; thus, heavy modeling work can be avoided. To achieve a pure, vibration-free response, we can apply the following shaped input \cite{kotaniemi2025data}:
\begin{equation}
\dot{x}^{*}_{d}(t) = \frac{k_{0}}{1+k_{0}} \, \dot{x}_{d}(t - T/2) 
+ \frac{1}{1+k_{0}} \, \dot{x}_{d}(t), 
\label{eq:shaper}
\end{equation}
where $\dot{x}_{d}$ is the desired velocity reference and 
$\dot{x}^{*}_{d}$ is the shaped velocity commands to the leader robot. Here, $T$ is the 
vibration period and $k_{0}$ is the damping factor representing the reduction in vibration magnitude over a time interval of $T/2$. We can express this
as
\begin{equation}
- k_{0} \, y(t - T/2) = y(t),\label{eq:ISS}
\end{equation}
where $y$ denotes the undesired vibration. Equation \eqref{eq:ISS} represents the physical principle of vibration cancellation arising from the periodic oscillatory nature of the signal, where the cancelling term is introduced with a delay of T/2.This input shaping can be easily implemented using the block diagram shown in \autoref{fig:IS}. An important design consideration is that the shaping coefficients must satisfy $k_{1} + k_{2} = 1$, ensuring that no loss occurs in the velocity reference.
The two main parameters of an input shaper are the time delay $T/2$ and the damping factor $k_{0}$. These parameters can be determined after one movement of the payload using optical trackers, see section \ref{sec:SimExp}. A good tuning would, in theory, cancel the vibrations entirely. But in practice, finding the exact values is very difficult, since the system response depends on many external factors, such as delays in the control system. Selecting values of the sufficiently close parameters still leads to a significant, though not complete, reduction in vibrations.

\subsection{Force estimator}\label{sub:Fest}
The force estimator (see Figure \ref{fig:block}) ensures that the virtual force used as input to the admittance controller is a smoother representation of the interaction force at the follower manipulator, which in turn enables the controller to drive the pose error between the robots toward zero, effectively restoring the estimated interaction force to its initial value (zero in this case).
The gain $K_e$ of the force estimator is identified experimentally. The flexible body is rigidly attached to both robots, and only the leader is commanded to move. Then, the ratio of the measured force at the tip of the follower and the corresponding position error $\Delta x$ is evaluated as $K_e = F / \Delta x$. For sufficiently small displacements, this ratio $K_e$ remains approximately constant, and its value is adopted as the estimator gain. A force estimator provides a force signal with lower noise than the force sensor attached to the follower robot. 

\subsection{ Follower control architecture}
An admittance controller governs the follower robot to maintain compliance with the leader's motion. The admittance controller receives the interaction force provided by the force estimator (section \ref{sub:Fest}) and it generates a velocity command for the follower robot that makes it follow the leader, i.e., the follower robot is compliant to the leader robot.

The admittance controller can be defined as $\frac{\dot{x}_2}{F_{est}}=1/(Ms+B)$, where only the virtual mass $M$ and damping $B$ are considered, see \cite{keemink2018admittance}. As a result, the follower does not resist the leader’s motion through any elastic restoring force, but rather responds according to the selected inertial and damping parameters. The leader dictates the motion via the interaction force $F_{est}$, and the follower adapts through the admittance law and contributes to vibration suppression by being passive. Note that the admittance pole is at $\omega_a = B/M$. One should select the values of $B$ and $M$ that make $\omega_a$ smaller than the lowest natural frequency at which the payload vibrates to suppress vibrations. Also, the selection of $B$ and $M$ ($\omega_a $) should take into account the velocity control-loop bandwidth of the follower robot. To accurately track the velocity commands, the value of $\omega_a$ should not be bigger than the velocity control-loop bandwidth.

If the velocity control bandwidth of the follower robot is known, the maximum velocity $\dot{x}_{2_{max}}$ that the follower can track can be estimated. And, if the maximum interaction force $F_{est_{max}}$ is known, the value of $B$ can be estimated as $B\approx\frac{F_{est_{max}}}{\dot{x}_{2_{max}}}$, and $M$ can be computed as $M = B/\omega_a$.





\section{Simulation and Experiments}\label{sec:SimExp}

The collaborative method was tested in a simulation using Matlab/Simulink software. The Simscape tool was used for simulating the flexible structure using its multibody blocks. The robot leader and follower were simulated using transfer function blocks containing the dynamics of the robot's velocity controllers. The elastic model was implemented as in Figure \ref{fig:block} to estimate the force. The values of the admittance parameters, and input shaping gains used in simulation are $M=0.01$~[kg], $B=6$~[Ns/m], and $K=0$~[N/m], and $K_1=0.49$ and $K_2=0.51$, respectively.

The experimental setup is composed of a KUKA Quantek robot (leader) lifting a flexible acrylic sheet from one side, and an iiwa robot (follower) holding the other side of the flexible body. An NDI Vega XT position sensor was used to measure the vibrations of the flexible body. Six markers were attached along the flexible beam, with the first and last markers defining a reference segment. The vibration amplitude was computed as the sum of the distances of the intermediate markers from this reference segment. Measurements were recorded both during and after the motion to capture the transient and residual vibrations. The sensor provides data at a sampling frequency of up to 60 Hz. The experimental test was performed using the admittance parameters $M=0.03$~[kg], $B=3$~[Ns/m], and $K=0$~[N/m], and input shaping gains $K_1=0.25$ and $K_2=0.75$


The testing scenario is to move the payload 10 cm in 0.5 seconds, a considerably fast and challenging motion that likely induces more vibrations. The scenario was tested in simulations and with the real robots using four different control strategies, recording the payload vibrations.
\begin{figure}
  \begin{center}
    \includegraphics[angle=0, width=7cm]{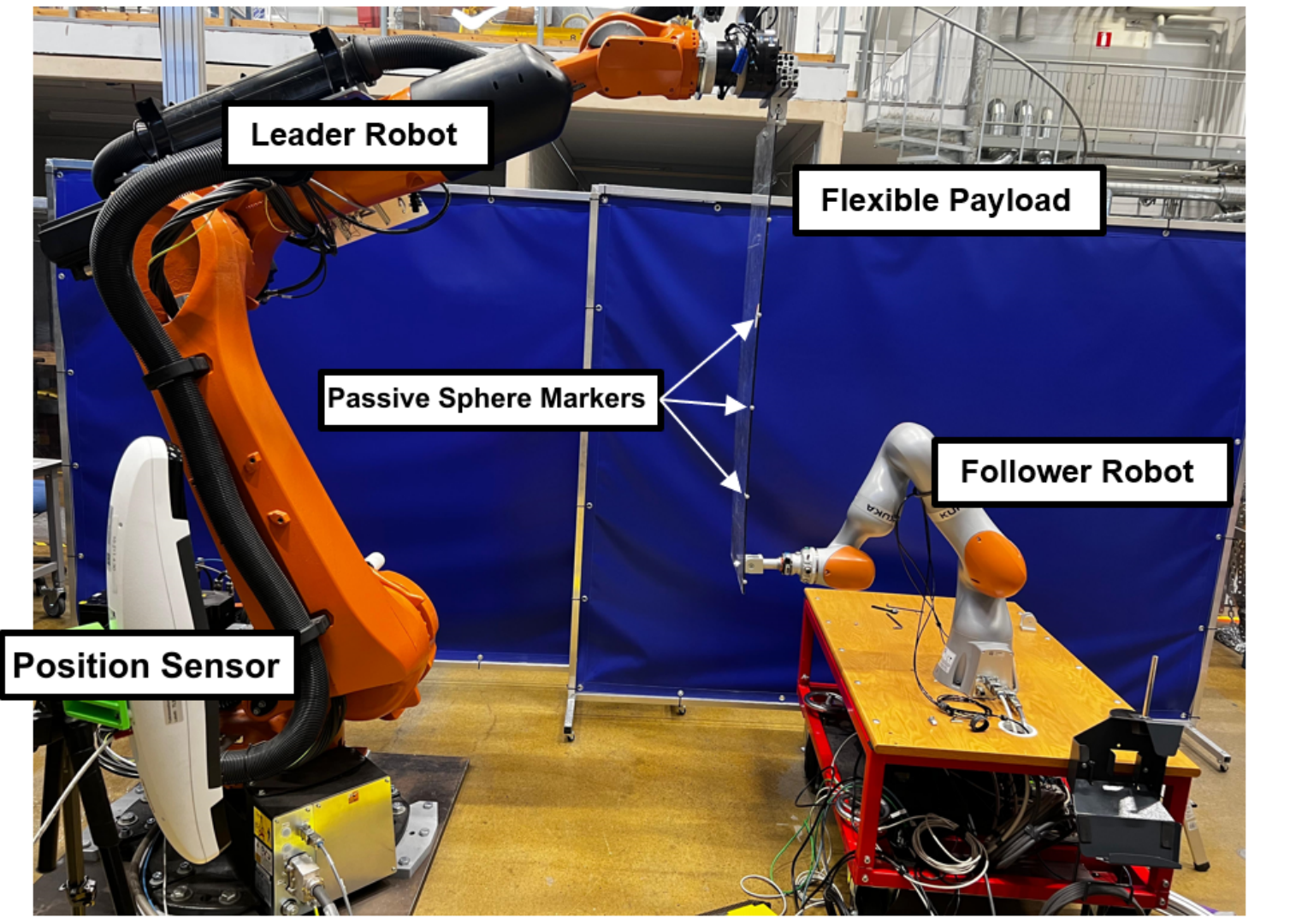}
  \end{center}
    \caption[]{Experimental setup}
    \label{fig:fig5}
\end{figure}

\begin{figure}
  \begin{center}
    \includegraphics[angle=0, width=9cm]{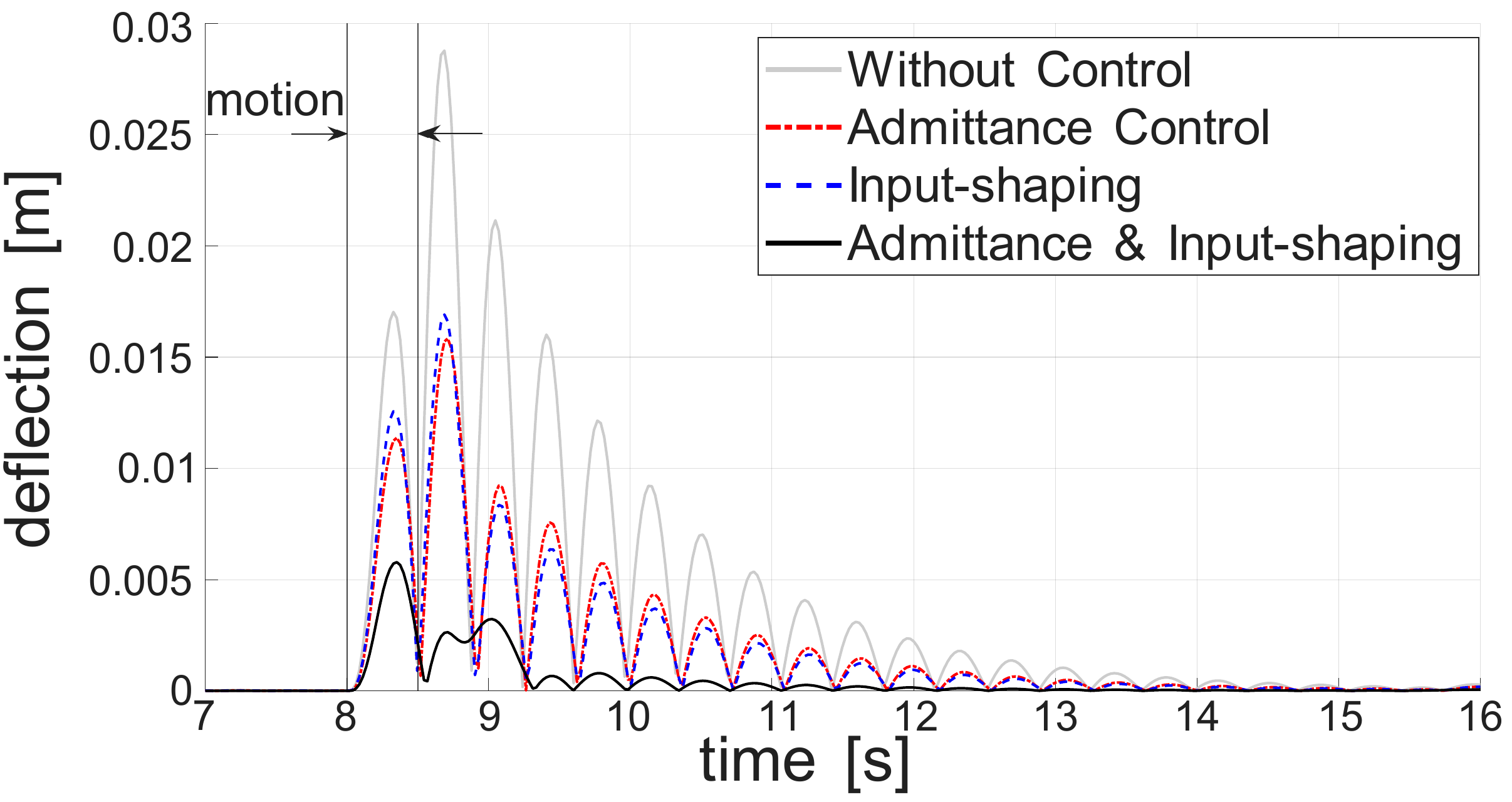}
  \end{center}
    \caption[]{ Simulation results of the vibrations of the flexible beam over time with different control methods}
    \label{fig:fig7}
\end{figure}

\begin{figure}
  \begin{center}
    \includegraphics[angle=0, width=9cm]{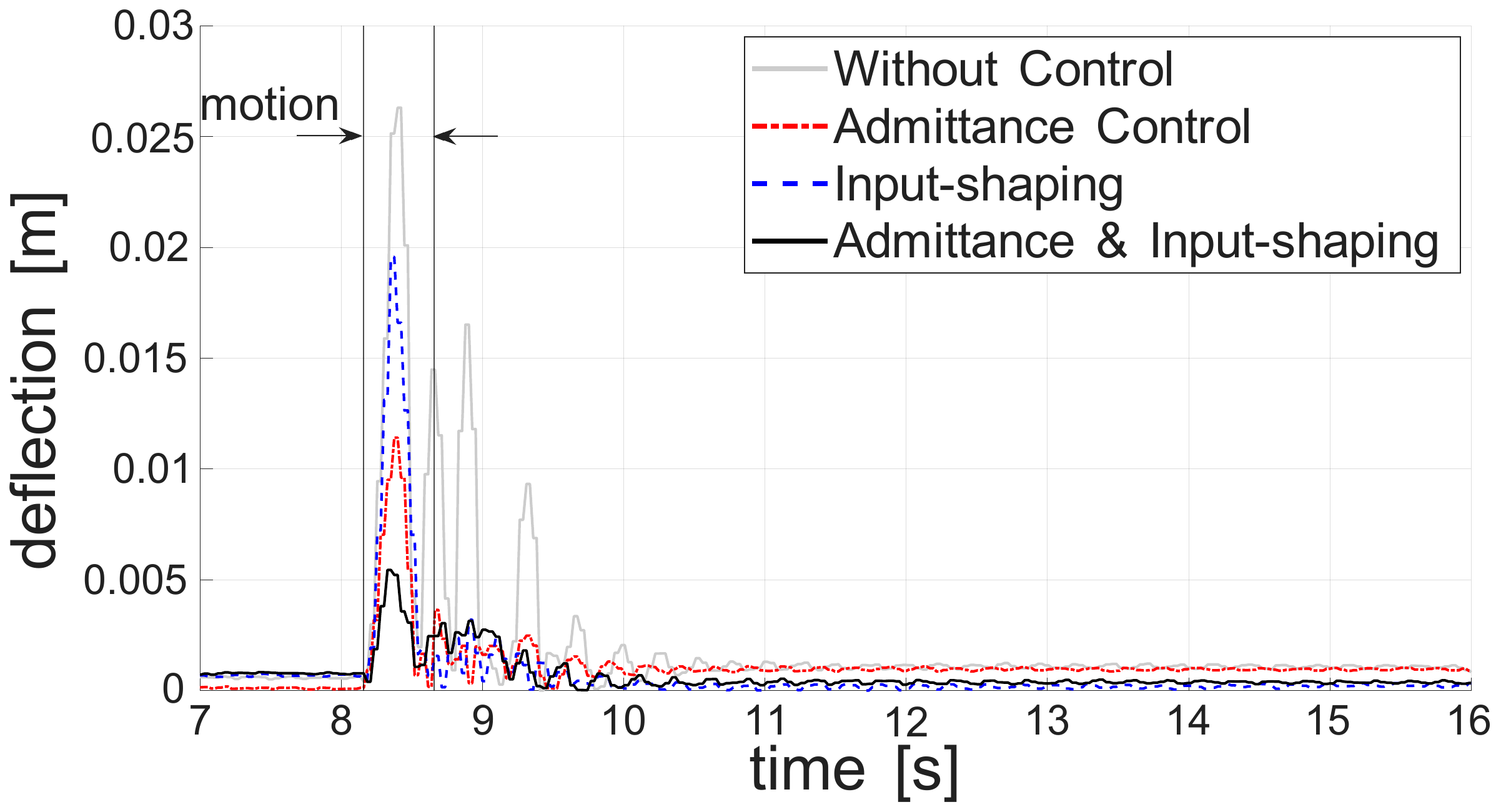}
  \end{center}
    \caption[]{ Experimental results of the vibrations of the flexible beam over time with different control methods}
    \label{fig:fig8}
\end{figure}

Figs.~\ref{fig:fig7} and \ref{fig:fig8} show the vibrations recorded from simulations and experiments, respectively. The vertical black lines in the figures denote the time window of the transient motion. The four control strategies are compared in Table~\ref{tab:control_comparison} using metrics such as the highest amplitude (during transient motion), settling time, and amplitude reduction percentage (compared with no control case). The percentage of reduction is defined as
\[
\text{Amp. red. (\%)} = \frac{A_{\text{none}} - A_{\text{control}}}{A_{\text{none}}} \times 100,
\]
where \(A\) denotes the peak vibration amplitude.
 \begin{table}[t]
\centering
\caption{Comparison of control strategies in simulation and experiment.}
\setlength{\tabcolsep}{3pt}
\begin{tabular}{lp{2cm}ccc}
\toprule
\shortstack{Case \\ {  }} & \shortstack{Control type\\ {  }} &  \shortstack{Highest amp. \\ {[mm]}}&  \shortstack{Settling time \\ {[s]}} &  \shortstack{Amp. red. \\ {[\%]}}\\
\midrule
\multirow{4}{*}{Simulation}
& None               & 28.8 & 4.65 & --  \\
& Admittance         & 15.6   & 3.57   & 45.8  \\
& Input-shaping      & 16.5 & 3.57   & 42.7  \\
& Adm.\,+\,IS        & 5.8 & 1.26   & 79.9  \\
\midrule
\multirow{4}{*}{Experiment}
& None               & 26.3 & 2.38  & --  \\
& Admittance         & 11.4 & 1.37   & 56.7  \\
& Input-shaping      & 19.6 & 1.2   & 25.5  \\
& Adm.\,+\,IS        & 5.4 & 1.18   & 79.5  \\
\bottomrule
\end{tabular}
\label{tab:control_comparison}
\end{table}

From Figs.~\ref{fig:fig7} and \ref{fig:fig8}, and Table \ref{tab:control_comparison}, one can observe that the response shows slowly decaying vibrations without control. Adding control clearly improves both peak amplitude and settling time. In simulation, admittance provides slightly better suppression than input shaping. In the laboratory, input-shaping is less effective at reducing peak amplitude (only \(25.5\%\)) but still yields faster settling than admittance control. This is because it avoids exciting the mode —an effect that becomes evident after about \(T/2\) of the vibration period— while admittance control introduces damping from the beginning of the motion. Pairing input shaping with admittance yields the most effective suppression of residual vibrations, achieving roughly \(\approx 80\%\) peak reduction. Differences between simulation and experiment arise from unmodeled dynamics and sensor/actuator limits.

\section{Conclusion}\label{sec:Conclu}
In this work, we present an effective solution for suppressing vibrations during the collaborative handling of a flexible payload by two heterogeneous manipulators. We achieved this by using an admittance controller in conjunction with a force estimator and an input shaper. This dissipates potential energy (vibrations) and reduces excitation of the dominant mode. Across simulation and lab trials, the combined strategy provided the largest vibration decay. 

The proposed approach aims to provide a solution for the manipulation of massive and large objects, such as in-vessel remote handling of breeding blankets. For future work, a 3D-printed model of the breeding blanket will be tested as the flexible payload, and various obstacle scenarios can be tested. The extension of our method to three dimensions is expected. Therefore, separate admittance parameters are required for each linear axis and for each rotation axis, depending on the system's minimum velocities.

Our current method uses industrial robots controlled in Cartesian space. Then, the application of our method in larger systems controlled in Cartesian space, such as cranes, is possible.


\bibliographystyle{IEEEtran}
\bibliography{refs}

\end{document}